\documentclass[10pt,twocolumn,letterpaper]{article}

\usepackage{cvpr} 
\usepackage{times}
\usepackage{epsfig}
\usepackage{graphicx}
\usepackage{amsmath}
\usepackage{amssymb}

\usepackage{hyperref}
\usepackage{multirow} 
\usepackage[table]{xcolor}
\usepackage{booktabs}
\usepackage{url}

\begin{document}

\title{AG-VPReID.VIR: Bridging Aerial and Ground Platforms for \\
Video-based Visible-Infrared Person Re-ID}

\author{Huy Nguyen, Kien Nguyen, Akila Pemasiri, Akmal Jahan, Clinton Fookes, and Sridha Sridharan\\
School of Electrical Engineering and Robotics, Queensland University of Technology\\
\small \{nguyet91, k.nguyenthanh, a.thondilege, akmal.jahan, c.fookes, s.sridharan\}@qut.edu.au
}


\maketitle
\thispagestyle{empty}

\begin{abstract}
Person re-identification (Re-ID) across visible and infrared modalities is crucial for 24-hour surveillance systems, but existing datasets primarily focus on ground-level perspectives. While ground-based IR systems offer nighttime capabilities, they suffer from occlusions, limited coverage, and vulnerability to obstructions—problems that aerial perspectives uniquely solve. To address these limitations, we introduce AG-VPReID.VIR, the first aerial-ground cross-modality video-based person Re-ID dataset. This dataset captures 1,837 identities across 4,861 tracklets (124,855 frames) using both UAV-mounted and fixed CCTV cameras in RGB and infrared modalities. AG-VPReID.VIR presents unique challenges including cross-viewpoint variations, modality discrepancies, and temporal dynamics. Additionally, we propose TCC-VPReID, a novel three-stream architecture designed to address the joint challenges of cross-platform and cross-modality person Re-ID. Our approach bridges the domain gaps between aerial-ground perspectives and RGB-IR modalities, through style-robust feature learning, memory-based cross-view adaptation, and intermediary-guided temporal modeling. Experiments show that AG-VPReID.VIR presents distinctive challenges compared to existing datasets, with our TCC-VPReID framework achieving significant performance gains across multiple evaluation protocols. Dataset and code are available at
\href{https://github.com/agvpreid25/AG-VPReID.VIR}{https://github.com/agvpreid25/AG-VPReID.VIR}.
\end{abstract}


\section{Introduction}
\label{sec:intro}
Person re-identification (Re-ID) aims to match images or videos of individuals captured across different cameras, times, or viewpoints. It has become a crucial task in computer vision with applications in surveillance and security systems~\cite{BeyondIntra, Ye2021DeepLF}. Traditional Re-ID methods focus on matching RGB images obtained from ground-based cameras under well-lit conditions~\cite{Liu2020PairbasedUA, Li2019DomainAwareUC}. However, real-world surveillance scenarios often involve varying illumination conditions, including low-light or nighttime environments, where RGB cameras may fail to capture useful information~\cite{wu2017rgb, lin2022learning}. In such cases, infrared (IR) imaging provides a viable solution due to its ability to capture images in low-light conditions~\cite{dai2018cross,liao2015person}.

\begin{figure}[!t] 
    \centering
    \includegraphics[width=1\linewidth]{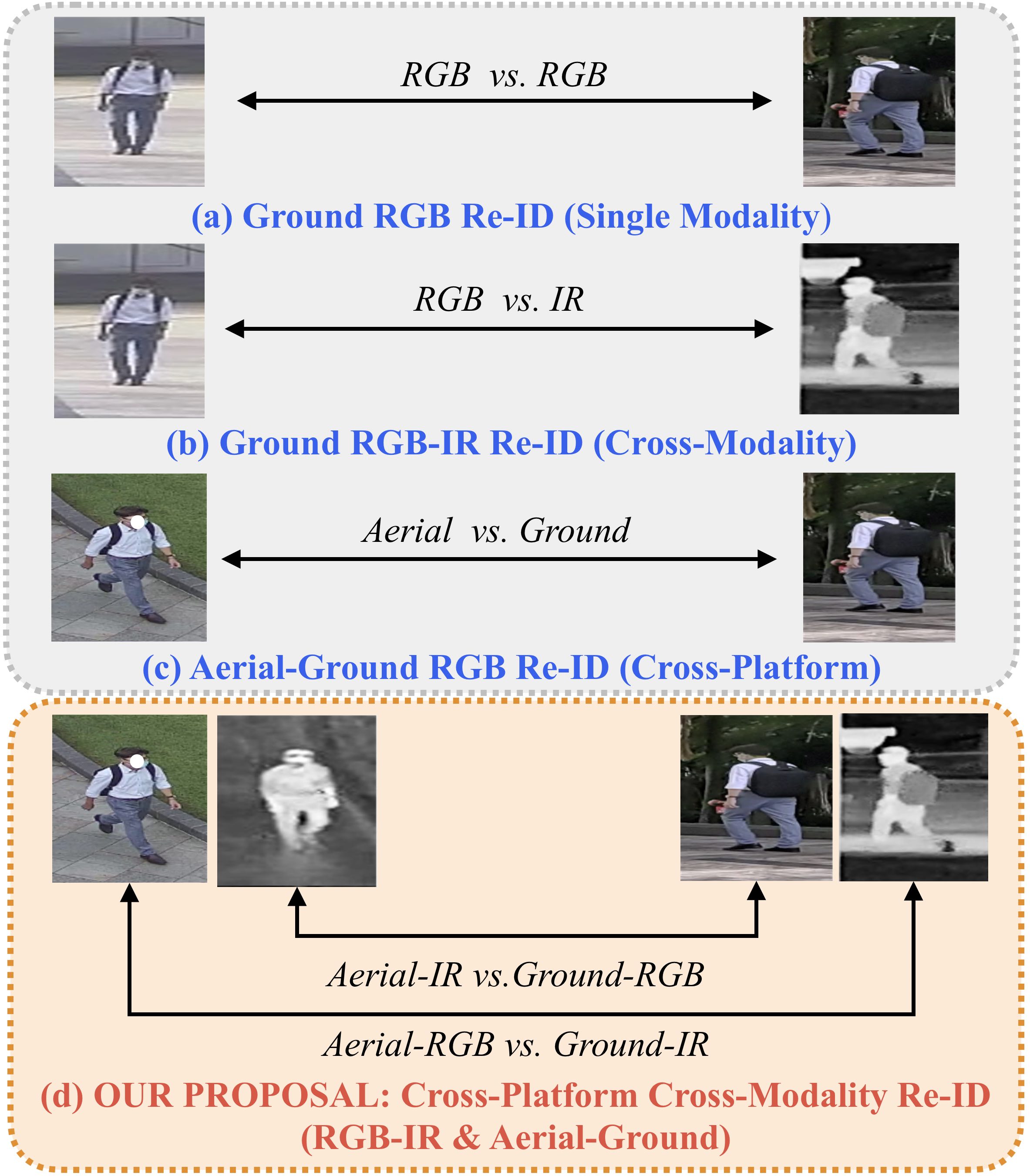}
   \caption{Existing person Re-ID works mainly focus on ground-based setting, either (a) single-modality RGB vs. RGB or (b) cross-modality RGB vs. IR. Recent advances in aerial-ground Re-ID only operate in (c) single-modality RGB vs. RGB. We propose a new cross-platform cross-modality setting (d) for person Re-ID.}
    \label{fig:proposed_dataset}
\end{figure}

Within person Re-ID, several works have explored the aerial-ground setting \cite{nguyen2023aerial, nguyen2023ag}. Nguyen et al. \cite{nguyen2024ag} introduced AG-ReID.v2 with 100,502 RGB images of 1,615 identities, while Zhang et al. \cite{zhang2024cross} proposed G2A-VReID with 185,907 video frames from 2,788 identities. However, both operate exclusively in RGB modality. Despite significant progress in cross-modality Re-ID for ground-based cameras, the combination of aerial-ground perspectives with RGB-IR modalities remains unexplored.

Cross-modality person Re-ID, particularly between RGB and IR images, has attracted significant interest for enabling 24-hour surveillance by bridging the day-night gap~\cite{wu2017rgb, lin2022learning}. Several datasets have been proposed for cross-modality Re-ID, such as SYSU-MM01~\cite{wu2017rgb}, RegDB~\cite{nguyen2017person}, HITSZ-VCM~\cite{lin2022learning}, and BUPTCampus~\cite{YunhaoDu2024}. However, these datasets primarily focus on ground-based surveillance and do not consider the cross-modality scenario for aerial-ground setting, which is critical for comprehensive surveillance systems involving both UAVs and ground cameras.

To address these limitations, we introduce AG-VPReID.VIR, a novel aerial-ground cross-modality video-based person Re-ID dataset. This dataset extends existing datasets by incorporating both RGB and IR modalities captured from aerial and ground perspectives. The combined challenges of cross platforms (aerial vs. ground) and cross modalities (RGB vs. IR) presents unique challenges to be addressed. The dataset contains 4,861 tracklets from 1,837 identities, captured by multiple UAV-mounted and fixed CCTV cameras in both modalities. This diverse collection presents unique challenges including cross-viewpoint variations, modality discrepancies, and temporal dynamics, making it a valuable benchmark for developing robust Re-ID systems. Fig.~\ref{fig:proposed_dataset} illustrates our proposed dataset and its differences compared to previously proposed settings. 

Furthermore, we propose TCC-VPReID (Three-stream architecture for Cross-platform Cross-modality Video-based Person Re-ID), a novel architecture specifically designed for RGB-IR aerial-ground person Re-ID. Our approach aims to effectively leverage both spatial and temporal information while addressing the significant domain gaps between aerial-ground views and RGB-IR modalities. Through extensive experiments on AG-VPReID.VIR, we demonstrate that our method achieves superior performance compared to existing state-of-the-art approaches.

The main contributions of this work are: (1) We introduce the first aerial-ground cross-modality video-based person Re-ID dataset, enabling research into more practical surveillance systems; (2) We establish baseline performance metrics through comprehensive benchmarking of state-of-the-art approaches; (3) We develop TCC-VPReID, a three-stream architecture that is purposely designed to address aerial-ground viewpoint variations and RGB-IR modality discrepancies through style-robust feature learning, memory-based cross-view adaptation, and temporal modeling. The dataset, baseline models, and our approach will be made publicly available.

\section{Prior Work}
\label{sec:prior}

\begin{figure}[!htbp]
    \centering
    \includegraphics[width=0.65\linewidth]{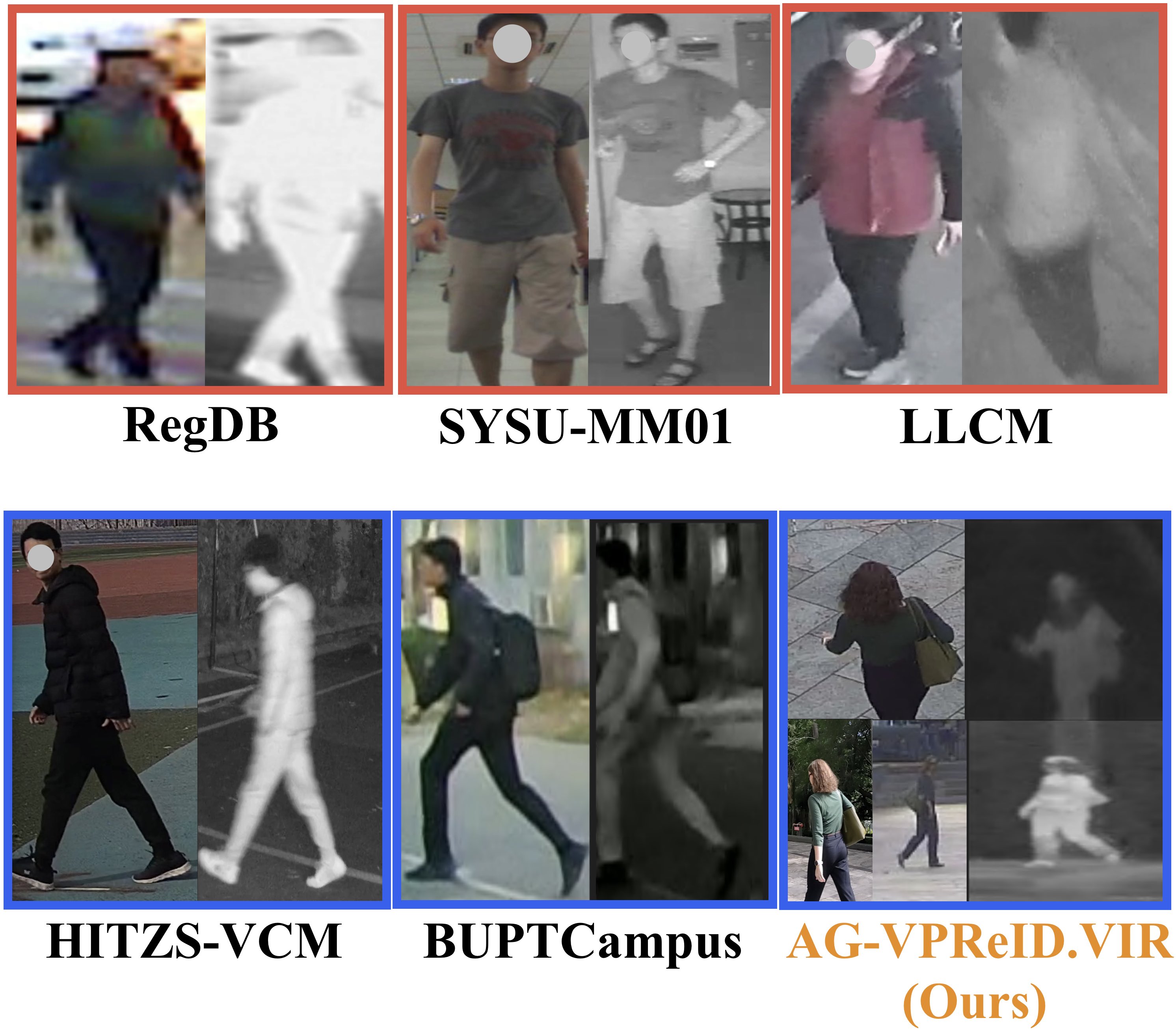}
    \caption{Our visible-infrared images vs. existing datasets' visible-infrared images. {Red border} indicates image-based vs. {blue border} indicates video-based setting. }
    \label{fig:dataset_compared}
\end{figure}

\noindent\textbf{Visible-Infrared Person Re-ID Datasets:} 
RGB-IR person re-identification has gained interest for 24-hour surveillance. SYSU-MM01 \cite{wu2017rgb} pioneered with 491 identities across 6 cameras, while RegDB \cite{nguyen2017person} provided 412 identities with paired RGB-thermal images. Video-based datasets emerged with HITSZ-VCM \cite{lin2022learning} (927 identities, 463,000 frames) and BUPTCampus \cite{YunhaoDu2024} (3,080 identities, 1.87 million frames). However, these datasets focus solely on ground-level perspectives, lacking the aerial viewpoints crucial for comprehensive surveillance. Table~\ref{tab:dataset_comparison} and Fig.~\ref{fig:dataset_compared} compare our proposed dataset with other visible-infrared datasets.

\begin{table}[!t]
    \centering
    \caption{Comparison of AG-VPReID.VIR with other person re-identification datasets.}
    \label{tab:dataset_comparison}
    \resizebox{0.48\textwidth}{!}{
    \begin{tabular}{l|cccc|ccc}
    \hline
    \multirow{2}{*}{Dataset} & \multicolumn{4}{c|}{View \& Modality} & \multicolumn{3}{c}{Statistics} \\
    \cline{2-8} 
    & \multicolumn{2}{c}{Ground} & \multicolumn{2}{c|}{Aerial} & \multirow{2}{*}{\#IDs} & \multirow{2}{*}{\#Tracklets} & \multirow{2}{*}{\#Frames}  \\ 
    & RGB & IR & RGB & IR & & &  \\
    \hline
    \multicolumn{8}{l}{\textbf{Image-based Datasets}} \\
    \hline
    AG-ReID.v2 \cite{nguyen2024ag} & \checkmark & - & \checkmark & - & 1,615 & - & 100,502 \\
    RegDB \cite{nguyen2017person} & \checkmark & \checkmark & - & - & 412 & - & 8,240  \\
    SYSU-MM01 \cite{wu2017rgb} & \checkmark & \checkmark & - & - & 491 & - & 303,420  \\
    LLCM \cite{zhang2023diverse} & \checkmark & \checkmark & - & - & 1,064 & - & 46,767  \\
    \hline
    \multicolumn{8}{l}{\textbf{Video-based Datasets}} \\
    \hline
    G2A-VReID \cite{zhang2024cross} & \checkmark & - & \checkmark & - & 2,788 & 5,576 & 180,000 \\
    HITZS-VCM \cite{lin2022learning} & \checkmark & \checkmark & - & - & 927 & 21,863 & 463,259  \\
    BUPTCampus \cite{YunhaoDu2024} & \checkmark & \checkmark & - & - & 3,080 & 16,826 & 1,869,366  \\
    \hline
    \rowcolor{gray!15}
    \textbf{AG-VPReID.VIR (Ours)} & \checkmark & \checkmark & \checkmark & \checkmark & 1,837 & 4,861 & 124,855 \\
    \hline
    \end{tabular}
    }
\end{table}

\noindent\textbf{Aerial-Ground Person Re-ID Datasets:} 
%
%
Aerial-ground perspectives in Re-ID research emerged with AG-ReID.v2 \cite{nguyen2024ag} (100,502 RGB images, 1,615 identities) and G2A-VReID \cite{zhang2024cross} (185,907 video frames, 2,788 identities). However, these datasets only include visible spectrum imagery, limiting nighttime utility. 
Our dataset uniquely combines aerial and ground views in both modalities for robust Re-ID across varied lighting and viewpoints.

\noindent\textbf{Visible-Infrared Person Re-ID:}
Cross-modality person Re-ID methods fall into two main approaches: learning modality-shared features through specialized architectures \cite{ye2018hierarchical,ye2020cross}, and employing adversarial learning \cite{dai2018cross,wang2019learning} to align feature distributions between modalities. Most existing methods assume identical data distributions between training and testing phases—an assumption that does not hold in practical Re-ID scenarios with non-overlapping identity sets. Additionally, these methods primarily consider images from similar ground-level perspectives, overlooking significant viewpoint variations. Our work specifically addresses the joint challenges of cross-modality and cross-platform variations, which represent a significant gap in existing literature.

\noindent\textbf{Video-based Person Re-ID:} 
Video-based person Re-ID methods extract temporal information through three primary approaches: recurrent neural networks \cite{mclaughlin2016recurrent,xu2017jointly,liu2019spatial} for sequential dependency modeling, temporal pooling operations \cite{xu2017jointly,chung2017two,wu2018exploit} for frame-level feature aggregation, and 3D convolutions \cite{li2019multi,gu2020appearance,liu2019spatial} for direct spatio-temporal pattern extraction. Recent advances include spatial-temporal memory networks \cite{eom2021video,lin2022learning,YunhaoDu2024} that store representations of spatial distractors and temporal patterns. While these methods have shown success in single-domain settings, they typically address either cross-modality challenges \cite{lin2022learning,YunhaoDu2024} or cross-viewpoint variations \cite{zhang2024cross} in isolation. Our work aims to overcome these unprecedented domain gaps through a novel architecture optimized for aerial-ground RGB-IR video-based Re-ID.

\section{AG-VPReID.VIR dataset}
\label{sec:dataset}

\subsection{Dataset Collection} 
\label{data_collect}

The AG-VPReID.VIR dataset was collected at a university campus using a combination of UAVs, stationary CCTV cameras, and wearable devices. Inspired by a previously published dataset~\cite{nguyen2024ag}, which included RGB images from UAVs, CCTV, and wearable cameras, AG-VPReID.VIR enhances data diversity by integrating IR images captured from UAVs and CCTV cameras equipped with thermal sensors. This addition allows for comprehensive surveillance capabilities under various lighting conditions, particularly in low-light or nighttime environments where RGB cameras are less effective.

For the UAV platform, we employed DJI M600 Pro drones equipped with dual-sensor cameras capable of capturing synchronized RGB and IR images. The UAVs were operated at altitudes ranging from 15 to 45 meters to capture individuals from elevated perspectives with varying resolutions and scales.
For ground-level surveillance, we utilized two CCTV cameras, comprising a standard visible-light (RGB) camera and a thermal imaging (IR) camera.
The wearable devices provide ground-level RGB images from first-person viewpoints. Table \ref{tab:camera_specs} provides detailed specifications of our camera setup. Camera locations and view coverage are presented in the Appendix Sec.~\ref{sec:dataset_collection_map}.

The data collection was conducted over five months, comprising 10 distinct sessions. 
During each session, UAV flights were performed at multiple altitudes (15m, 25m, and 45m), with each flight lasting 15 minutes to ensure comprehensive coverage. We gathered a total of 124,855 images of 1,837 unique identities. The AG-VPReID.VIR dataset includes both RGB and IR images captured across different camera modalities and perspectives. The inclusion of IR imagery significantly enhances the dataset's applicability for person re-identification tasks under varied environmental conditions.

\begin{table}[ht]
    \centering
    \caption{Camera specifications used in data collection.}
    \resizebox{0.9\columnwidth}{!}{
    \label{tab:camera_specs}
    \begin{tabular}{lcccccc}
        \toprule
        \textbf{Device} & \textbf{Mod.} & \textbf{Brand} & \textbf{Model} & \textbf{Res.} & \textbf{FPS} & \textbf{Alt.}   \\ 
        \midrule
        CCTV & RGB & Bosch & N/A & $704{\times}480$ & $15$ & ${\approx}3$m \\
        Wearable & RGB & Vuzix & M4000 & 4K & $30$ & ${\approx}1.5$m \\ 
        UAV & RGB & DJI & XT2 & $3840{\times}2160$ & $30$ & $15{\sim}45$m \\
        UAV & NIR & DJI & XT2 & $640{\times}512$ & $30$ & $15{\sim}45$m \\
        CCTV & NIR & Avigilon & N/A & $800{\times}600$ & $30$ & ${\approx}4$m \\
        \bottomrule
    \end{tabular}
    }
\end{table}

\subsection{Labeling Process}
\label{label_process}
To ensure accurate cross-modal and cross-platform person re-identification, each individual in the dataset was manually annotated with a unique identity label consistent across RGB and IR images, as well as across UAV, CCTV, and wearable camera views. Initially, the YOLOv10x (Ultralytics) detector with pretrained weights as the backbone \cite{Jocher2024} were employed for person detection and tracking, extracting frames at regular intervals. Annotators manually reviewed the automatically detected tracks to correct any inaccuracies, ensuring that each tracklet corresponds to a single individual across all modalities and platforms. This two-stage process ensures high-quality annotations while maintaining efficiency in the labeling workflow.

\begin{figure}
    \centering
    \includegraphics[width=1\linewidth]{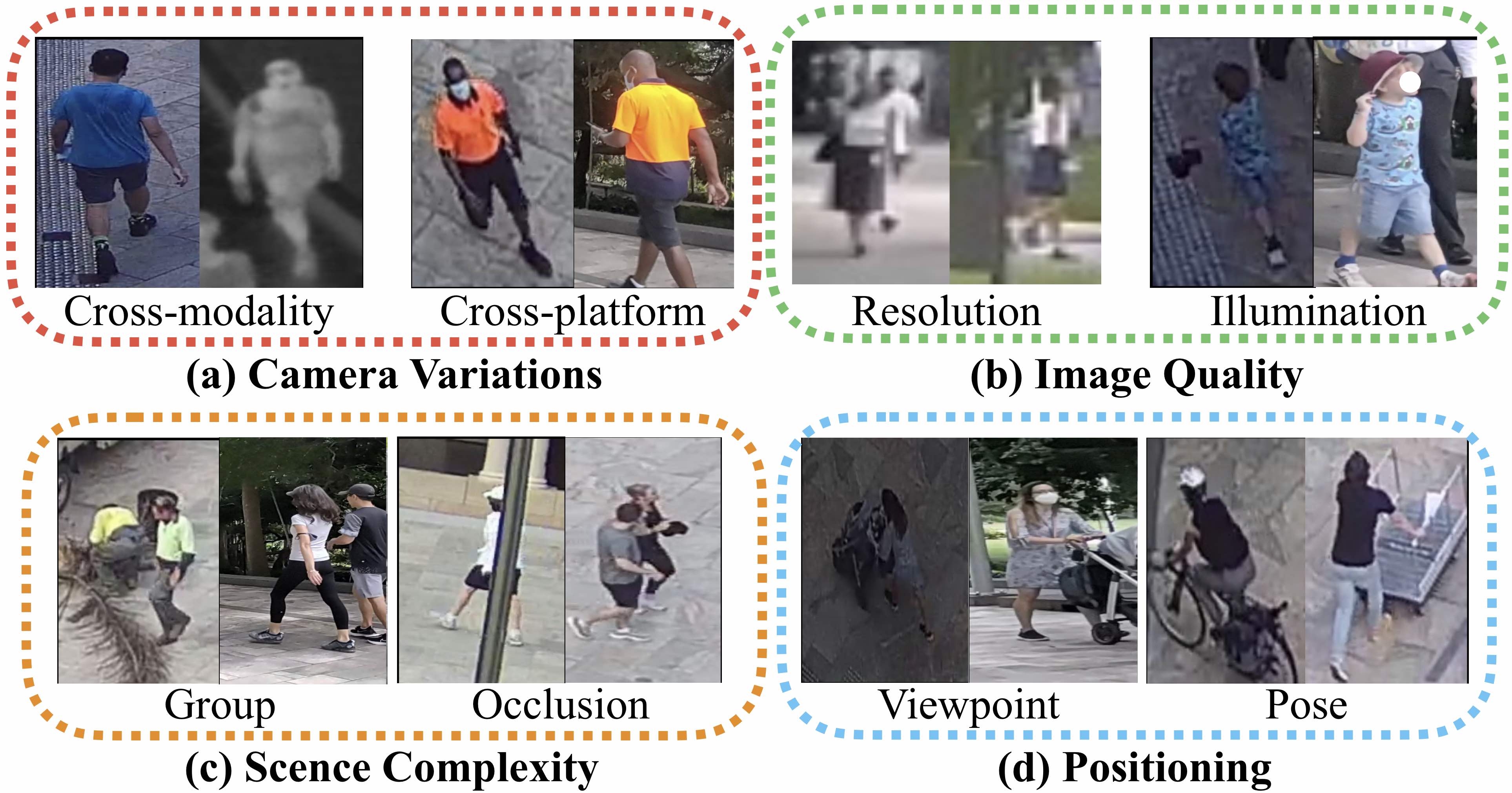}
    \caption{Challenges in our proposed dataset. }
    \label{fig:Challenges}
\end{figure}

The final dataset contains 124,855 images corresponding to 1,837 unique identities captured across 5 camera views. The labeling process accounted for challenges such as varying image resolutions, significant viewpoint differences, and modality discrepancies between RGB and IR images. Multiple annotators cross-verified the labels to enhance annotation accuracy. The dataset distribution analysis is presented in Appendix Sec.~\ref{sec:dataset_distribution}.

\subsection{Dataset Characteristics} 
\label{sec:data_chars}
The AG-VPReID.VIR dataset exhibits several unique characteristics compared to existing RGB-IR person re-identification datasets such as RegDB~\cite{nguyen2017person}, SYSU-MM01~\cite{wu2017rgb}, HITSZ-VCM~\cite{lin2022learning}, and BUPTCampus~\cite{YunhaoDu2024}. Table~\ref{tab:dataset_comparison} provides a comparative overview of these datasets. The key characteristics of the AG-VPReID.VIR dataset include:
\begin{itemize}
    \item \textbf{Unique Multi-Platform Multi-Modal Integration}: Unlike existing datasets, which either capture only ground-based RGB-IR data (SYSU-MM01, RegDB, HITSZ-VCM, BUPTCampus) or only RGB data across aerial-ground platforms (AG-ReID.v2, G2A-VReID), our dataset uniquely incorporates both RGB and IR modalities from both aerial and ground-based cameras, providing the first comprehensive cross-platform cross-modality collection for person Re-ID.
    \item \textbf{Unique Aerial IR}: Our dataset is the first to introduce IR images from UAV platforms for person ReID. As shown in Fig.~\ref{fig:Challenges}(a), aerial IR images are very challenging compared to Ground IR images, posing interesting challenges for the research community to address.
    \item \textbf{Novel Viewpoint and Scale Challenges}: The dataset captures individuals from various angles and distances, resulting in significant variations in appearance and scale. Due to different camera heights and types, UAV-captured images range from 31$\times$59 to 371$\times$678 pixels, while ground-based images span from 22$\times$23 to 172$\times$413 pixels, as illustrated in Fig.~\ref{fig:Challenges}(a) and Fig.~\ref{fig:Challenges}(b).
    \item \textbf{Other Challenges:} In addition, our dataset poses various challenges existing in person Re-ID including scene complexity with partially occluded subjects and individuals walking in groups as depicted in Fig.~\ref{fig:Challenges}(c), and positioning challenges from different viewpoints and poses as presented in Fig.~\ref{fig:Challenges}(d).
\end{itemize}


\subsection{Ethics and Privacy} 
\label{sec:ethics}
This study was conducted with full ethical approval from the institutional review board. All participants provided informed consent prior to data collection. To protect privacy, we employed the ``Deface" \cite{deface} for facial anonymization, implemented secure data storage protocols with restricted access, and established clear guidelines for responsible data usage.


\begin{figure*}[t]
    \centering
    \begin{tabular}{c}
        \includegraphics[width=0.9\linewidth]{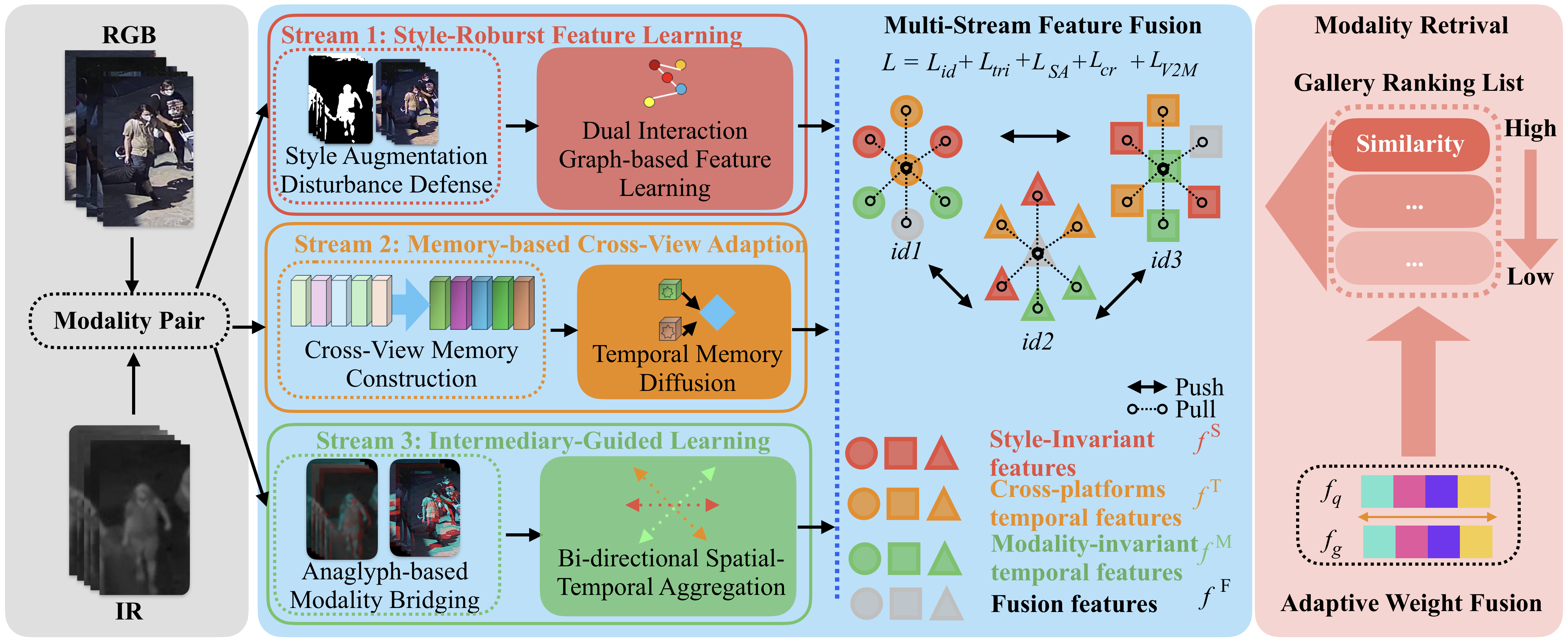} \\
    \end{tabular}
   \caption{Overview of our proposed three-stream architecture: (1) Style-Robust Stream for handling appearance variations via style augmentation and graph learning; (2) Memory-based Cross-View Stream for bridging aerial-ground perspectives; and (3) Intermediary-Guided Temporal Stream for addressing RGB-IR modality gaps. These streams combine in a fusion module for effective cross-platform, cross-modality person re-identification.}
    \label{fig:method}
\end{figure*}

\section{Our Method}
\label{sec:method}

We propose TCC-VPReID, a novel three-stream framework purposely designed to address the unique challenges of cross-platform cross-modality person re-identification. In particular, our approach is designed with complementary techniques to handle \textbf{(1)} the style variations and appearance distortions across platforms (\textit{Stream 1}), \textbf{(2)} the temporal and cross-view challenges of video-based (\textit{Stream 2}) and \textbf{(3)} the modality gap challenges between visible and infrared domains (\textit{Stream 3}). Fig.~\ref{fig:method} illustrates our architecture, highlighting three specialized streams and the feature fusion module that preserves complementary information while reconciling contradictions across platforms and modalities. 

\subsection{Stream 1: Style-Robust Feature Learning}
\label{sec:style_robust_stream}
To overcome the dramatic style variations and appearance distortions across aerial-ground platforms and RGB-IR modalities, our first stream implements a novel style-robust feature extraction mechanism. While inspired by recent style augmentation techniques~\cite{zhou2023video}, we significantly extend this approach with platform-specific style transformations tailored to the unique characteristics of aerial and ground perspectives.

This stream generates diverse training representations by strategically varying input frame styles:
{\small
\begin{align}
I_t^{aug} &= [\alpha I_t^R, \beta I_t^G, \gamma I_t^B],\\
I_t^{ir,aug} &= \delta I_t^{ir},
\end{align}
}
where $I_t$ and $I_t^{ir}$ denote the $t$-th RGB and infrared frames, respectively. $I_t^R$, $I_t^G$, $I_t^B$ represent the red, green, and blue channels of $I_t$, and $\alpha$, $\beta$, $\gamma$, and $\delta$ are sampled from a uniform distribution $U(0.5, 1.5)$ as proposed in~\cite{zhou2023video}. 

Our innovation lies in developing a cross-platform defense mechanism that maintains feature consistency despite the extreme viewpoint and modality variations. We employ strategically placed intra-modal style attacks between network's convolutional blocks:
{\small
\begin{align}
\tilde{f}_{conv3}^{i,j} = \frac{\sigma(f_{conv3}^{k,l})}{\sigma(f_{conv3}^{i,j})} \cdot (f_{conv3}^{i,j} - \mu(f_{conv3}^{i,j})) + \mu(f_{conv3}^{k,l}),
\end{align}
}
where $f_{conv3}^{i,j}$ denotes the feature from the third convolutional block for the $i$-th identity and $j$-th frame, $f_{conv3}^{k,l}$ represents a feature from a different identity-frame pair ($k$,$l$) selected for disturbance, and $\sigma(\cdot)$ and $\mu(\cdot)$ represent the variance and mean operations on the input feature. 
This approach forces the model to develop viewpoint-invariant and modality-invariant representations through our specialized optimization objective:
{\small
\begin{align}
L_{SA} = L_{dis} + L_{con},
\end{align}
}
where $L_{dis}$ enforces identity consistency through cross-entropy loss and $L_{con}$ enforces feature consistency through the distance between original and attacked features.

\subsection{Stream 2: Memory-based Cross-View Adaptation}
\label{sec:memory_stream}
The dramatic geometric and perspective changes between aerial and ground viewpoints represent one of the most challenging aspects of cross-platform Re-ID. Our second stream introduces a novel memory-based architecture that explicitly models and bridges these viewpoint differences through tailored representation learning and adaptive cross-view alignment.

The centerpiece of our innovation is the development of specialized view-specific memory representations~\cite{zhang2023camera,zheng2022plausible,bertocco2021unsupervised} that separately model aerial and ground viewpoints:
{\small
\begin{align}
M_{y_i}^{aerial} &= \frac{1}{N_i^{aerial}}\sum_{v_a \in y_i^{aerial}} v_a,\\
M_{y_i}^{ground} &= \frac{1}{N_i^{ground}}\sum_{v_a \in y_i^{ground}} v_a,
\end{align}
}
where $M_{y_i}^{aerial}$ and $M_{y_i}^{ground}$ are the memory representations for identity $y_i$ from aerial and ground views respectively, $v_a$ represents a sequence-level feature vector extracted from the visual encoder, $N_i^{aerial}$ and $N_i^{ground}$ represent the number of aerial and ground sequences for identity $y_i$, respectively. 
Unlike traditional approaches that use a single global memory per identity, our dual-memory architecture explicitly captures the distinct characteristics of each viewing angle.

We further introduce a cross-platform attention mechanism that dynamically updates each memory type:

{\small
\begin{align}
M_{y_i}^{m'} &= \mathcal{P}_{y_i}^{m} + M_{y_i}^{m},
\end{align}
}
where $m \in \{aerial, ground\}$ denotes the platform type, $M_{y_i}^{m'}$ is the updated memory representation, and $\mathcal{P}_{y_i}^{m}$ is a platform-specific prompt generated via a dual-branch attention-based decoder that explicitly shares information across platforms while preserving platform-specific characteristics.

Our innovative video-to-memory contrastive loss optimizes these representations:
{\small
\begin{align}
L_{V2M}(y_i) = -\frac{1}{|P(y_i)|}\sum_{p\in P(y_i)}\log\frac{\exp(v_p \cdot M_{y_i}^{m'}/\tau)}{\sum_{j=1}^{B}\exp(v_j \cdot M_{y_i}^{m'}/\tau)},
\end{align}
}
where $P(y_i)$ represents all positive samples for identity $y_i$ in the batch, $v_p$ is the feature vector of a positive sample, $v_j$ refers to the feature vector of the $j$-th sample in the batch, $\tau$ is a temperature parameter, and $B$ is the batch size.


\subsection{Stream 3: Intermediary-Guided Temporal Learning}
\label{sec:intermediary_stream}

The fundamental domain gap between visible and infrared modalities presents unique challenges, particularly when coupled with aerial-ground viewpoint variations. Our third stream introduces a novel intermediary-guided temporal learning module specifically designed to bridge these modality differences while preserving critical identity information.

Our core approach generates modality-invariant anaglyph representations via edge detection ~\cite{li2023intermediary}:

{\small
\begin{align}
a(i,j) = \sum_m \sum_n x(i+m, j+n)A(m,n) + k,
\end{align}
}
where $a(i,j)$ is the $(i,j)$-th pixel of anaglyph image $a$,  $x(i,j)$ represents the original image (RGB or IR), $A(m,n)$ denotes an edge detection operator with indices $m,n \in \{-1,0,1\}$, and $k$ is an offset value.

Our innovation lies in the development of a specialized 3D cross-attention mechanism that strengthens modality-invariant information while preserving spatial-temporal patterns critical for aerial-ground alignment. This is complemented by our cross-reconstruction constraint that explicitly minimizes the modality gap:
{\small
\begin{align}
L_{cr} = \sum_{t=1}^{b \times k} \|a_t^v - R(a_t^{ir})\|_2 + \sum_{t=1}^{b \times k} \|a_t^{ir} - R(a_t^v)\|_2,
\end{align}
}
where $a_t^v$ and $a_t^{ir}$ represent anaglyph features from visible and infrared data respectively, $R$ denotes a reconstruction network consisting of convolutional layers, $b$ is the mini-batch size, and $k$ is the number of frames in each video clip. 
This approach forces the model to learn modality-invariant features that are robust to the dramatic appearance differences between RGB and IR data.




\subsection{Multi-Stream Integration and Loss Functions}
\label{sec:integration}

The three streams provide complementary information for cross-platform cross-modality person re-identification. We integrate these streams through a feature fusion module and optimize using a joint loss function:
{\small
\begin{align}
L_{total} = L_{id} + \lambda_1 L_{tri} + \lambda_2 L_{SA} + \lambda_3 L_{cr} + \lambda_4 L_{V2M},
\end{align}
}
where hyperparameters $\lambda_1$, $\lambda_2$, $\lambda_3$, and $\lambda_4$ balance the contribution of each loss term from the three streams described earlier.  During inference, the fused feature serves as the final person representation for cross-domain matching, with adaptive weighting applied to each stream's contribution based on validation performance. 

\section{Experimental Results}
\label{sec:experiments}

\subsection{Datasets and Evaluation Metrics}
\label{sec:eval}
For a thorough evaluation, we validate our approach on two established visible-infrared benchmarks HITSZ-VCM \cite{lin2022learning} and BUPTCampus \cite{YunhaoDu2024}. We also conduct comprehensive experiments on our AG-VPReID.VIR dataset. As shown in Table~\ref{tab:dataset_stats}, we maintain a strict separation between training and testing identities, with 326 IDs (978 tracklets, 24,793 frames) reserved for training and the remaining identities used for testing across four challenging evaluation scenarios: ground-to-ground, aerial-to-aerial, ground-to-aerial, and aerial-to-ground. Each protocol includes both visible-to-infrared (V2I, where visible images serve as queries and IR images as gallery) and infrared-to-visible (I2V, where IR images serve as queries and visible images as gallery) matching directions. For I2V experiments, we further increase the difficulty by introducing additional distractor identities in the gallery set, creating a more realistic surveillance scenario with 1,184 distractor IDs contributing 30,532-34,687 frames depending on the platform. We adopt standard person Re-ID evaluation metrics: Cumulative Matching Characteristic (CMC) along with mean Average Precision (mAP) \cite{YunhaoDu2024, yu2024tf} to comprehensively measure retrieval performance across all ranking positions.

\begin{table}[!htbp]
    \centering
    \renewcommand{\arraystretch}{1.2}
    \setlength{\tabcolsep}{4pt}
    \caption{Train/test statistics for AG-VPReID.VIR. Q=Query, G=Gallery. V2I: visible to infrared, I2V: infrared to visible.}
    \label{tab:dataset_stats}
    \resizebox{\columnwidth}{!}{
    \begin{tabular}{l|l|c|c|c|c}
    \toprule
    \textbf{Set} & \textbf{Platforms} & \textbf{Modalities} & \textbf{\# IDs} & \textbf{\# Tracklets (Q/G)} & \textbf{\# Frames (Q/G)} \\
    \midrule
    \midrule
    Training & All & -- & 326 & 978 & 24,793 \\
    \midrule
    \multirow{8}{*}{Testing} & \multirow{2}{*}{Ground to Ground} & V2I & 199 & 313/199 & 9,357/3,018 \\
    &  & I2V* & 199 & 199/313 & 3,018/9,357 \\
    \cmidrule{2-6}
    & \multirow{2}{*}{Aerial to Aerial} & V2I & 174 & 174/174 & 6,284/3,115 \\
    & & I2V* & 174 & 174/174 & 3,115/6,284 \\
    \cmidrule{2-6}
    & \multirow{2}{*}{Ground to Aerial} & V2I & 116 & 184/116 & 5,360/1,701 \\
    &  & I2V* & 260 & 260/260 & 3,530/10,939 \\
    \cmidrule{2-6}
    & \multirow{2}{*}{Aerial to Ground} & V2I & 260 & 260/260 & 10,939/3,530 \\
    &  & I2V* & 116 & 116/184 & 1,701/5,360 \\
    \bottomrule
    \multicolumn{6}{l}{* For all I2V experiments, 1,184 additional distractor IDs are added to the gallery set.}
    \end{tabular}
    }
    \end{table}

\begin{table*}[ht]
\centering
\caption{Comprehensive Comparison of Cross-Modal Person Re-Identification Methods.}
\label{tab:comprehensive_comparison}
\resizebox{0.775\textwidth}{!}{
\begin{tabular}{l|c|cc|cc|cc|cc|cc|cc}
\toprule
& & \multicolumn{8}{c|}{\textbf{Ground $\leftrightarrow$ Ground}} & \multicolumn{4}{c}{\textbf{Ground $\leftrightarrow$ Ground}} \\
\cline{3-14}
& & \multicolumn{4}{c|}{\textbf{HITSZ-VCM}} & \multicolumn{4}{c|}{\textbf{BUPTCampus}} & \multicolumn{4}{c}{\textbf{AG-VPReID.VIR}} \\
\cline{3-14}
\multirow{2}{*}{\textbf{Method}} & \multirow{2}{*}{\textbf{Venue}} & \multicolumn{2}{c|}{\textbf{I2V}} & \multicolumn{2}{c|}{\textbf{V2I}} & \multicolumn{2}{c|}{\textbf{I2V}} & \multicolumn{2}{c|}{\textbf{V2I}} & \multicolumn{2}{c|}{\textbf{I2V}} & \multicolumn{2}{c}{\textbf{V2I}} \\
\cline{3-14}
& & \textbf{R1} & \textbf{mAP} & \textbf{R1} & \textbf{mAP} & \textbf{R1} & \textbf{mAP} & \textbf{R1} & \textbf{mAP} & \textbf{R1} & \textbf{mAP} & \textbf{R1} & \textbf{mAP} \\
\midrule
\midrule
\rowcolor{gray!10} \multicolumn{14}{l}{\textbf{Image-based Methods}} \\
AlignGAN~\cite{wang2019rgb} & ICCV'19 & 42.15 & 27.43 & 44.62 & 29.73 & 27.99 & 30.32 & 35.37 & 35.13 & 4.12 & 6.32 & 3.94 & 8.21 \\
LbA~\cite{park2021learning} & ICCV'21 & 46.38 & 30.69 & 49.30 & 32.38 & 32.09 & 32.93 & 39.07 & 37.06 & 5.23 & 7.14 & 4.82 & 9.43 \\
MPANet~\cite{wu2021discover} & CVPR'21 & 46.51 & 35.26 & 50.32 & 37.80 & 33.45 & 34.17 & 40.56 & 38.22 & 5.94 & 8.21 & 5.63 & 10.55 \\
SPOT~\cite{chen2022structure} & TIP'22 & 52.23 & 37.10 & 53.67 & 38.96 & 37.52 & 38.24 & 44.12 & 41.03 & 7.32 & 9.87 & 6.42 & 11.63 \\
HCT~\cite{liu2020parameter} & TMM'20 & 55.39 & 38.62 & 57.25 & 39.73 & 38.92 & 39.47 & 45.62 & 42.18 & 7.83 & 10.54 & 7.21 & 12.37 \\
DDAG~\cite{ye2020dynamic} & ECCV'20 & 54.62 & 39.26 & 59.03 & 41.50 & 40.44 & 40.86 & 46.30 & 43.05 & 8.12 & 11.25 & 7.83 & 13.42 \\
MMN~\cite{zhang2021towards} & MM'21 & 53.24 & 39.82 & 56.43 & 41.47 & 40.86 & 41.71 & 43.70 & 42.80 & 7.95 & 11.08 & 7.42 & 13.15 \\
CAJL~\cite{ye2021channel} & ICCV'21 & 56.59 & 41.49 & 60.13 & 42.81 & 40.49 & 41.46 & 45.00 & 43.61 & 9.12 & 12.86 & 8.76 & 15.32 \\
VSD~\cite{tian2021farewell} & CVPR'21 & 54.53 & 41.18 & 57.52 & 43.45 & 41.27 & 42.05 & 47.84 & 44.63 & 8.45 & 12.03 & 8.12 & 14.27 \\
AGW~\cite{Ye2021DeepLF} & TPAMI'21 & 47.23 & 36.12 & 51.85 & 38.50 & 36.38 & 37.36 & 43.70 & 41.10 & 6.74 & 9.23 & 6.17 & 11.08 \\
DEEN~\cite{zhang2023diverse} & CVPR'23 & 65.32 & 50.16 & 68.42 & 50.83 & 49.81 & 48.59 & 53.70 & 50.43 & 11.23 & 16.45 & 10.54 & 19.23 \\
SGIEL~\cite{feng2023shape} & CVPR'23 & 67.65 & 52.30 & 70.23 & 52.54 & 48.32 & 47.15 & 51.42 & 49.27 & 12.43 & 17.92 & 11.21 & 20.45 \\
\midrule
\rowcolor{gray!10} \multicolumn{14}{l}{\textbf{Video-based Methods}} \\
TCLNet~\cite{hou2020temporal} & ECCV'20 & 48.32 & 36.45 & 52.13 & 38.52 & 37.15 & 36.87 & 41.32 & 38.92 & 9.23 & 12.54 & 8.37 & 15.76 \\
CAViT~\cite{wu2022cavit} & ECCV'22 & 54.83 & 40.26 & 58.18 & 41.63 & 42.63 & 41.87 & 46.52 & 43.15 & 10.45 & 14.83 & 9.72 & 17.63 \\
DART~\cite{yang2022learning} & CVPR'22 & 62.85 & 44.26 & 63.92 & 46.43 & 52.43 & 49.10 & 53.33 & 50.45 & 11.56 & 15.82 & 10.23 & 18.76 \\
MITML~\cite{lin2022learning}  & CVPR'22 & 63.74 & 45.31 & 64.54 & 47.69 & 49.07 & 47.50 & 50.19 & 46.28 & 12.16 & 16.93 & 10.87 & 19.42 \\
AuxNet~\cite{YunhaoDu2024} & TIFS'23 & 51.05 & 45.99 & 54.58 & 48.70 & 66.48 & 64.11 & 65.23 & 62.19 & 15.70 & 16.46 & 12.89 & 22.57 \\
CST\cite{feng2024cross} & TMM'24 & 69.44 & 51.16 & 72.64 & 51.16 & 54.92 & 52.87 & 56.35 & 53.26 & 14.23 & 19.37 & 12.64 & 22.05 \\
IBAN~\cite{li2023intermediary} & TCSVT'23 & 65.03 & 48.77 & 69.58 & 50.96 & 53.42 & 51.36 & 54.17 & 52.43 & 6.03 & 9.46 & 3.83 & 10.16 \\
SAADG~\cite{zhou2023video} & ACM MM'23 & 69.22 & 53.77 & 73.13 & 56.09 & 55.76 & 53.42 & 57.83 & 54.82 & 13.07 & 18.75 & 11.82 & 21.14 \\
\rowcolor{blue!10} \textbf{Ours} & - & \textbf{72.16} & \textbf{56.43} & \textbf{75.92} & \textbf{57.84} & \textbf{69.73} & \textbf{67.25} & \textbf{68.47} & \textbf{65.38} & \textbf{36.18} & \textbf{41.56} & \textbf{46.33} & \textbf{59.23} \\
\bottomrule
\end{tabular}
}
\end{table*}

\begin{table*}[ht]
    \centering
    \caption{Comprehensive Results on Cross-Platforms Cross-Modal AG-VPReID.VIR Dataset.}
    \label{tab:comprehensive_results}
    \renewcommand{\arraystretch}{1.3}
    \setlength{\tabcolsep}{4pt}
    \resizebox{0.8\textwidth}{!}{
    \begin{tabular}{l|ccccc|ccccc|ccccc|ccccc}
    \toprule
    & \multicolumn{10}{c|}{\textbf{Aerial $\boldsymbol{\rightarrow}$ Ground}} & \multicolumn{10}{c}{\textbf{Ground $\boldsymbol{\rightarrow}$ Aerial}} \\
    \cmidrule(lr){2-11} \cmidrule(lr){12-21}
    & \multicolumn{5}{c|}{\textbf{I2V}} & \multicolumn{5}{c|}{\textbf{V2I}} & \multicolumn{5}{c|}{\textbf{I2V}} & \multicolumn{5}{c}{\textbf{V2I}} \\
    \cmidrule(lr){2-6} \cmidrule(lr){7-11} \cmidrule(lr){12-16} \cmidrule(lr){17-21}
    \multirow{-3}{*}{\textbf{Method}} & \textbf{R1} & \textbf{R5} & \textbf{R10} & \textbf{R20} & \textbf{mAP} & \textbf{R1} & \textbf{R5} & \textbf{R10} & \textbf{R20} & \textbf{mAP} & \textbf{R1} & \textbf{R5} & \textbf{R10} & \textbf{R20} & \textbf{mAP} & \textbf{R1} & \textbf{R5} & \textbf{R10} & \textbf{R20} & \textbf{mAP} \\
    \midrule
    St1  & 7.76 & 17.24 & 22.41 & 35.34 & 9.54 & 7.31 & 23.46 & 39.62 & 52.31 & 17.46 & 12.31 & 26.54 & 40.00 & 47.69 & 20.64 & 14.13 & 27.72 & 44.02 & 57.61 & 23.05 \\
    St2 & 11.21 & 28.45 & 36.21 & 45.69 & 17.68 & 40.38 & 70.77 & 82.69 & 92.31 & 54.40 & 27.31 & 52.69 & 63.08 & 71.92 & 39.27 & 27.17 & 50.54 & 65.76 & 79.35 & 39.25 \\
    St3 & 2.59 & 11.21 & 17.24 & 22.41 & 4.86 & 5.38 & 13.46 & 22.69 & 36.92 & 11.50 & 3.08 & 14.23 & 18.46 & 26.92 & 9.10 & 7.07 & 20.65 & 33.15 & 47.28 & 15.61 \\
    \midrule
    St12 & 15.52 & 31.90 & 43.10 & 51.72 & 20.91 & \textbf{46.92} & \textbf{77.31} & \textbf{86.15} & \textbf{92.31} &\textbf{ 59.71} & {34.23} & {56.54} & {70.77} & {78.85} & {45.62} & 30.43 & 56.52 & 67.39 & 80.98 & 42.45 \\
    St13 & 9.48 & 18.97 & 28.45 & 37.93 & 10.94 & 8.08 & 26.15 & 41.92 & 56.92 & 17.96 & 13.85 & 28.46 & 38.08 & 48.46 & 21.43 & 14.67 & 32.07 & 45.65 & 59.24 & 24.11 \\
    St23 & 18.10 & \textbf{34.48} & 42.24 & 45.69 & 21.22 & 41.54 & 74.23 & 85.00 & 92.31 & 56.31 & 31.92 & 55.38 & 66.54 & 78.08 & 43.01 & 31.52 & 53.80 & 66.85 & 78.80 & 42.49 \\
    \rowcolor[gray]{0.95} St123 & \textbf{19.83} & {31.90} & \textbf{43.10} & \textbf{51.72} & \textbf{22.61} & {46.54} & 76.54 & 85.38 & {91.92} & {59.69} & \textbf{34.23} & \textbf{56.54} & \textbf{70.77} & \textbf{78.85} & \textbf{45.62} & \textbf{31.52} & \textbf{57.07} & \textbf{67.39} & \textbf{80.98} & \textbf{42.92} \\
    \bottomrule
    \end{tabular}
    }
\end{table*}

\begin{table}[ht]
\centering
\caption{Aerial-to-Aerial Re-ID results in AG-VPReID.VIR.}
\label{tab:aerial_to_aerial_results}
\renewcommand{\arraystretch}{1.3}
\setlength{\tabcolsep}{4pt}
\resizebox{\columnwidth}{!}{
\begin{tabular}{l|ccccc|ccccc}
\toprule
& \multicolumn{5}{c|}{\textbf{RGB $\boldsymbol{\rightarrow}$ IR}} & \multicolumn{5}{c}{\textbf{IR $\boldsymbol{\rightarrow}$ RGB}} \\
\cmidrule(lr){2-6} \cmidrule(lr){7-11}
\multirow{-2}{*}{\textbf{Method}} & \textbf{R1} & \textbf{R5} & \textbf{R10} & \textbf{R20} & \textbf{mAP} & \textbf{R1} & \textbf{R5} & \textbf{R10} & \textbf{R20} & \textbf{mAP} \\
\midrule
St1  & 17.82 & 30.46 & 42.53 & 62.07 & 26.26 & 8.05 & 21.26 & 29.89 & 40.80 & 15.95 \\
St2  & 25.86 & 47.13 & 58.05 & 74.14 & 37.26 & 11.49 & 28.16 & 33.91 & 43.10 & 19.69 \\
St3 & 8.05 & 30.46 & 38.51 & 51.15 & 18.85 & 6.90 & 12.07 & 18.39 & 27.59 & 10.99 \\
\midrule
St12 & 30.46 & \textbf{56.90} & \textbf{66.09} & \textbf{77.01} & 42.40 & 20.69 & 39.66 & 50.57 & {56.90} & 29.89 \\
St13 & 20.11 & 38.51 & 50.57 & 62.07 & 30.43 & 12.64 & 24.71 & 32.18 & 43.10 & 19.28 \\
St23 & 31.61 & 54.02 & 63.22 & 75.86 & 43.17 & 20.11 & 31.61 & 43.10 & 49.43 & 26.93 \\
\rowcolor[gray]{0.95} St123 & \textbf{33.91} & 56.32 & 64.37 & 76.44 & \textbf{44.72} & \textbf{21.84} & \textbf{40.80} & \textbf{53.45} & \textbf{56.90} & \textbf{30.92} \\
\bottomrule
\end{tabular}
}
\end{table}

\subsection{Implementation Details} 
\label{sec:imp}
Our TCC-VPReID framework uses PyTorch on an NVIDIA A100 GPU. The three streams include: (1) Style-Robust stream with ResNet50 \cite{he2016deep} using style augmentation from $U(0.5, 1.5)$ and adaptive instance normalization; (2) Memory-based stream with CLIP ViT-B/16 \cite{radford2021learning} encoder and a 2-layer transformer decoder \cite{vaswani2017attention}; (3) Intermediary-Guided stream with a dual-branch architecture using anaglyph data and bidirectional LSTMs \cite{liu2019spatial}. These backbones and network architectures were selected based on the best performing models in the respective papers \cite{zhou2023video, yu2024tf, li2023intermediary}. Training uses Adam optimizer (initial lr=$3.5 \times 10^{-4}$) with cosine annealing. Each batch contains 8 identities × 4 sequences × 8 frames. Model trains for 120 epochs with hyperparameters $\lambda_1=1.0$, $\lambda_2=1.5$, $\lambda_3=1.0$, and $\lambda_4=1.5$ (see Appendix Sec.~\ref{sec:hyperparameter_ablation}).

\subsection{Comparison with State-of-the-Art Methods}
\label{sec:comp}
We evaluate our method against state-of-the-art approaches on HITSZ-VCM, BUPTCampus, and AG-VPReID.VIR datasets. We focus on ground-to-ground evaluation (Table~\ref{tab:comprehensive_comparison}) for fair comparison, with cross-platform ablations in Tables~\ref{tab:comprehensive_results} and \ref{tab:aerial_to_aerial_results}.

\vspace{3px}
\hspace{-12px}\textbf{Performance on Existing Datasets.}
On HITSZ-VCM, our method achieves 72.16\%/56.43\% (Rank-1/mAP) for I2V and 75.92\%/57.84\% for V2I, outperforming SAADG by 2.94\%/2.66\% and 2.79\%/1.75\%. On BUPTCampus, we reach 69.73\%/67.25\% for I2V and 68.47\%/65.38\% for V2I, exceeding AuxNet by 3.25\%/3.14\% and 3.24\%/3.19\%.

\begin{figure*}
    \centering
    \includegraphics[width=0.85\linewidth]{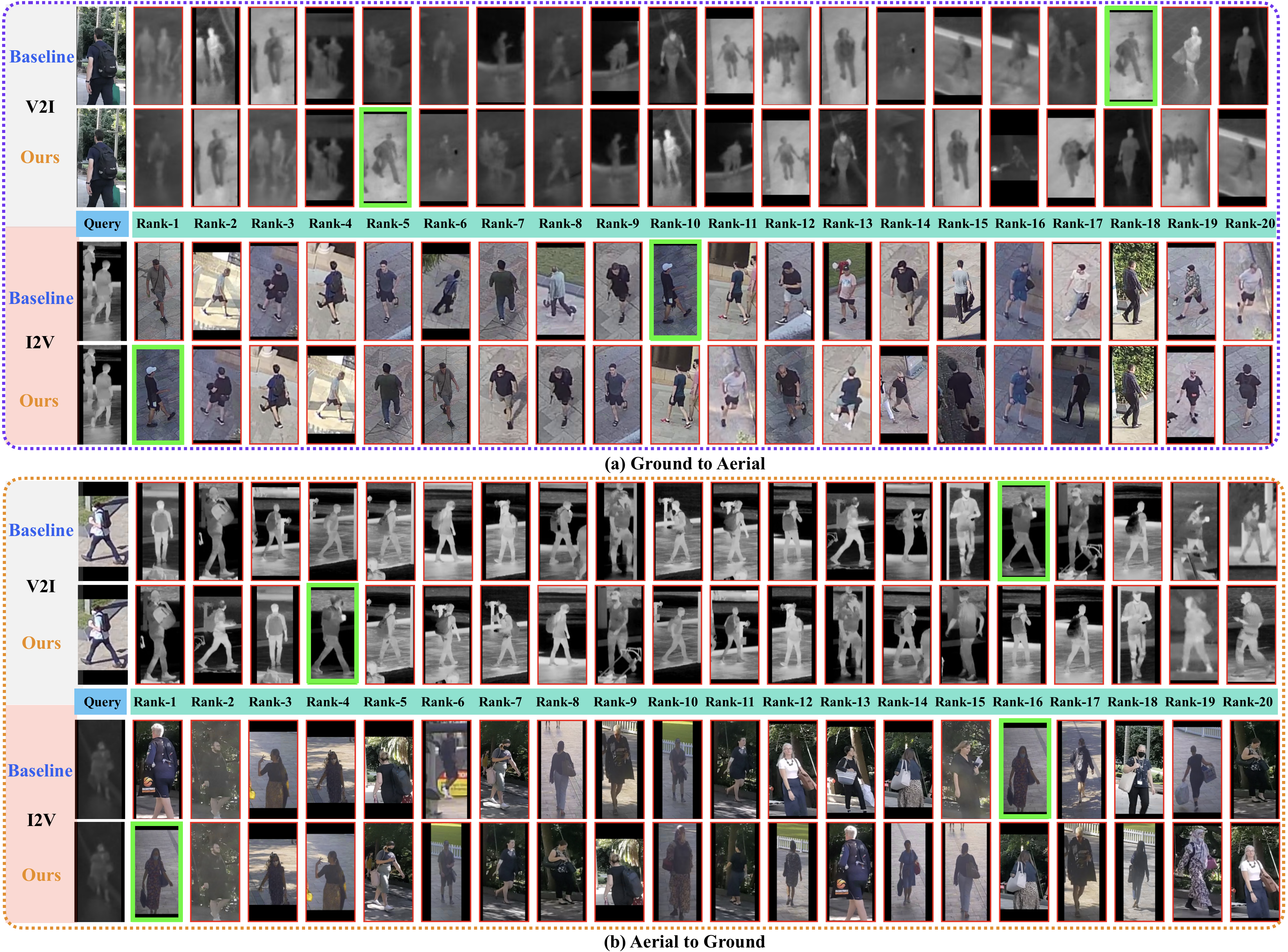}
    \caption{Visualization of top-20 ranking results on AG-VPReID.VIR dataset under aerial-ground cross-platform settings for I2V and V2I queries (I: infrared, V: visible). Red bounding boxes indicate incorrect matches, green indicate correct matches.}
    \label{fig:cross-platform-ranking}
\end{figure*}

\vspace{3px}
\hspace{-12px}\textbf{Performance on AG-VPReID.VIR.}
Our method shows the most significant gains on AG-VPReID.VIR, where we achieve 36.18\%/41.56\% for I2V and 46.33\%/59.23\% for V2I. These represent substantial improvements over previous best methods—SAADG (13.07\%/18.75\%), AuxNet (15.70\%/16.46\%), and CST (14.23\%/19.37\%)—with absolute gains of 20.48-23.11\% in Rank-1 and 22.19-25.10\% in mAP for I2V. The performance gap is even more pronounced for V2I, where our approach outperforms the best competitors by 33.69-35.11\% in Rank-1 and 37.18-39.07\% in mAP. Notably, all methods experience dramatic performance drops when transitioning to AG-VPReID.VIR's cross-platform setting. While SAADG's Rank-1 accuracy plummets from 69.22\% to 13.07\% (56.15\% relative drop) and MITML falls from 63.74\% to 12.16\% (51.58\% drop), our method maintains higher performance across datasets. 

\subsection{Ablation Study}
\label{sec:abl}
We analyze the contribution of each component in our TCC-VPReID framework. Streams are denoted as St1 (Style-Robust Feature Learning), St2 (Memory-based Cross-View Adaptation), and St3 (Intermediary-Guided Temporal Learning). Results are in Tab.~\ref{tab:comprehensive_results} and Tab.~\ref{tab:aerial_to_aerial_results}.

\vspace{1px}
\noindent\textbf{Individual Stream Performance.} 
St2 consistently outperforms other individual streams (11.21\% Rank-1 for Aerial$\rightarrow$Ground I2V, 40.38\% for V2I), demonstrating temporal memory's effectiveness for cross-domain challenges. St1 performs second-best, highlighting style-invariant features' importance, while St3 performs lowest individually.

\vspace{1px}
\noindent\textbf{Combined Streams.} 
The St12 combination significantly improves performance over individual streams (15.52\% Rank-1 for Aerial$\rightarrow$Ground I2V vs. 11.21\% for St2 alone). The full three-stream model (St123) achieves best overall results, confirming the complementary contributions of each stream. 

\vspace{1px}
\noindent\textbf{Modality Direction Asymmetry.} 
Performance significantly differs between V2I and I2V directions, with Aerial$\rightarrow$Ground showing 46.54\% vs. 19.83\% Rank-1. This asymmetry suggests infrared queries face greater challenges when searching visible galleries, likely due to information loss in thermal imagery.

\vspace{1px}
\noindent\textbf{Platform Comparison.} 
Same-platform cross-modality retrieval (33.91\% Rank-1 for Aerial$\rightarrow$Aerial RGB$\rightarrow$IR) outperforms cross-platform scenarios (19.83\% for Aerial$\rightarrow$Ground I2V), quantifying the challenge of cross-platform variations beyond modality differences.  The same observation can be made for Ground$\rightarrow$Ground. Additional cross-platform results detailed in Table~\ref{tab:cross_platform_comparison}.

\begin{table}[h]
\centering
\caption{Performance comparison on AG-VPReID.VIR ground-to-ground vs. aerial-to-ground scenarios (infrared-to-visible).}
\label{tab:cross_platform_comparison}
\scriptsize
\begin{tabular}{l|cc|cc}
\toprule
\textbf{Method} & \multicolumn{2}{c|}{\textbf{Ground$\rightarrow$Ground}} & \multicolumn{2}{c}{\textbf{Aerial$\rightarrow$Ground}} \\
 & \textbf{R1} & \textbf{mAP} & \textbf{R1} & \textbf{mAP} \\
\midrule
MITML [CVPR'22] & 12.16 & 16.93 & 3.95 & 6.45 \\
SAADG [ACM MM'23] & 13.07 & 18.75 & 4.21 & 7.83 \\
CST [TMM'24] & 14.23 & 19.37 & 4.87 & 8.12 \\
\rowcolor{gray!15}
\textbf{Ours} & \textbf{36.18} & \textbf{41.56} & \textbf{19.83} & \textbf{22.61} \\
\bottomrule
\end{tabular}
\end{table}

\subsection{Qualitative Results}
\label{sec:visualization}
Fig.~\ref{fig:cross-platform-ranking} compares our TCC-VPReID approach with the baseline~\cite{zhou2023video} across cross-platform cross-modality settings. Results show significant ranking improvements in all scenarios (\eg aerial-IR to ground-RGB from rank-16 to rank-1), confirming that our three-stream architecture effectively tackles variations in viewpoint and modality.

\section{Conclusion} 
\label{sec:conclusion}
We introduce AG-VPReID.VIR, the first comprehensive dataset for video-based cross-modality person re-identification across aerial and ground platforms, with 4,861 tracklets from 1,837 identities. Our proposed TCC-VPReID framework effectively tackles viewpoint variations and modality discrepancies through style-robust feature learning, memory-based cross-view adaptation, and intermediary-guided temporal learning. Experimental results show TCC-VPReID outperforms state-of-the-art methods by over 20\% in both Rank-1 accuracy and mAP, setting a new challenging benchmark for robust cross-platform surveillance systems.


{\small
\bibliographystyle{ieee}
\bibliography{egbib}
}

\clearpage
\setcounter{page}{1}
\maketitlesupplementary

\section{Dataset Collection}
\label{sec:dataset_collection_map}

\begin{figure}[!htbp]
    \centering
    \includegraphics[width=1\linewidth]{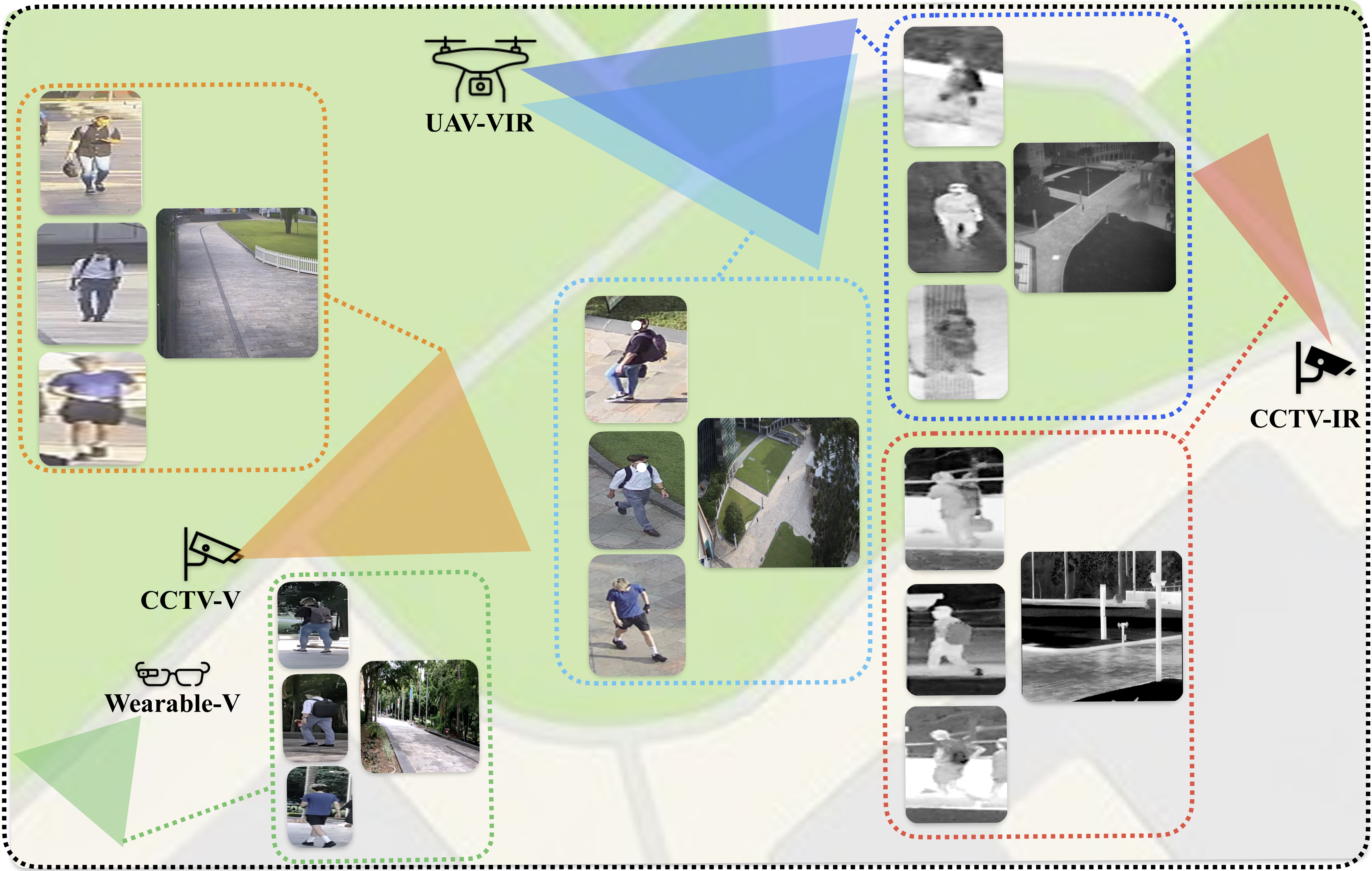}
    \caption{Data collection map. V denotes visible, IR means infrared, and VIR stands for both visible and infrared cameras.}
    \label{fig:data_collection_map}
\end{figure}

Fig.~\ref{fig:data_collection_map} illustrates the strategic camera placement used for our AG-VPReID.VIR dataset collection across a university campus. The map highlights the diverse sensing infrastructure deployed to capture comprehensive cross-platform and cross-modality data: visible-only cameras (V), infrared-only cameras (IR), and dual-modality cameras capturing visible and infrared modalities simultaneously (VIR). We strategically positioned these cameras to maximize coverage while ensuring minimal overlap between different sensing modalities and viewing angles. The UAVs were deployed at fixed hovering positions (marked on the map) but at varying altitudes (15m, 25m, and 45m) to capture diverse aerial perspectives, while ground-based CCTV and wearable cameras provided complementary viewpoints. This comprehensive sensing network enabled the collection of synchronized multi-view data across different environmental conditions, creating challenging scenarios with varying illumination, occlusion, and viewing angles that closely mirror real-world surveillance scenarios.

\section{Dataset Distribution Analysis}
\label{sec:dataset_distribution}
\begin{figure}[!htbp]
\centering
\includegraphics[width=0.75\linewidth]{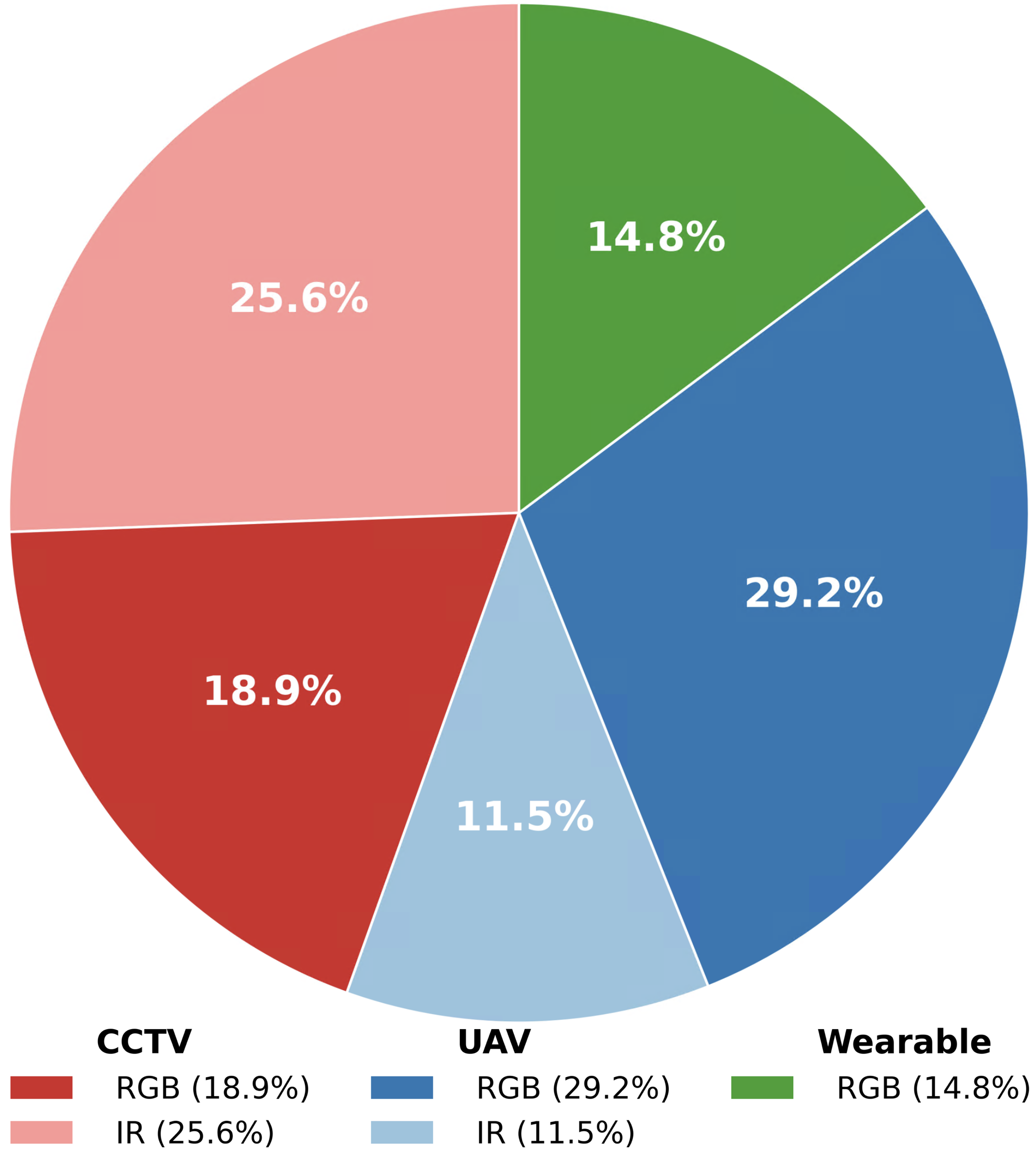}
\caption{Distribution of identity presence across different camera types and modalities (RGB and IR) in our AG-VPReID.VIR dataset. }
\label{fig:camera_id_distribution}
\end{figure}

Fig.~\ref{fig:camera_id_distribution} illustrates the distribution of data across different camera types and modalities in our AG-VPReID.VIR dataset. The distribution shows a deliberate balance between RGB and IR modalities while maintaining diversity across platforms. UAV RGB contributes the largest portion (29.2\%) of the dataset, providing rich aerial visible-light perspectives, while CCTV IR represents the second largest share (25.6\%), offering extensive ground-level thermal imagery. The complementary nature of CCTV RGB (18.9\%), UAV IR (11.5\%), and wearable RGB (14.8\%) ensures comprehensive coverage across all deployment scenarios. This balanced distribution is critical for training robust cross-modality cross-platform Re-ID models, as it provides sufficient examples for the algorithm to learn the distinctive characteristics of each viewpoint-modality combination, particularly the challenging aerial IR perspective that is unique to our dataset.

\begin{table*}[!htbp]
    \centering
    \small
    \caption{Overview of {TCC-VPReID}: A \textbf{T}hree-Stream Architecture for \textbf{C}ross-Platform \textbf{C}ross-Modality \textbf{V}ideo-based \textbf{P}erson Re-ID}
    \label{tab:model_overview}
    \resizebox{0.9\textwidth}{!}{
    \renewcommand{\arraystretch}{1.1}
    \begin{tabular}{p{0.20\textwidth}p{0.38\textwidth}p{0.35\textwidth}}
        \toprule
        \textbf{Stream} & \textbf{Challenges Addressed} & \textbf{Key Components} \\ 
        \midrule
        \textbf{Stream 1:} Style-Robust Feature Learning & 
        • Intra/inter-modal variations \newline
        • Frame-level style variations\newline
        • Appearance distortions across platforms &
        • Style Augmentation \& Disturbance Defense\newline
        • Cross-View \& Cross-Modal Graph Interaction\newline
        • Domain-Adversarial Alignment \\
        \midrule
        \textbf{Stream 2:} Temporal Memory Adaptation & 
        • Aerial-ground viewpoint variations\newline
        • Temporal misalignment in sequences\newline
        • Unstable motion patterns across platforms &
        • Cross-View Memory Construction\newline
        • Transformer-Based Video Encoder\newline
        • Temporal Memory Diffusion \\
        \midrule
        \textbf{Stream 3:} Modality-Invariant Representation & 
        • Modality gap (visible vs. infrared)\newline
        • Noisy frames in aerial/ground videos\newline
        • Information asymmetry between modalities &
        • Anaglyph-based Modality Bridging\newline
        • Bidirectional Spatial-Temporal Aggregation\newline
        • Modality Decoupling \\
        \midrule
        \textbf{Feature Fusion} & 
        • Preserving complementary information\newline
        • Reconciling cross-platform/modality features\newline
        • Balancing spatial, temporal, and modality cues &
        • Multi-level Feature Integration\newline
        • Cross-stream Knowledge Transfer\newline
        • Joint Optimization Framework \\
        \bottomrule
    \end{tabular}
    }
\end{table*}

\section{Challenges Addressed and Key Components of streams in the proposed architecture}
\label{sec:arch_overview}
Table~\ref{tab:model_overview} presents the comprehensive architecture of our TCC-VPReID framework, detailing how each stream addresses specific challenges in aerial-ground visible-infrared person re-identification. The three complementary streams work in concert to overcome the complex variations encountered in cross-platform cross-modality scenarios. 
Stream 1 focuses on style-robust feature learning to handle intra/inter-modal variations, frame-level style variations, and appearance distortions across platforms through components like style augmentation, cross-view graph interaction, and domain-adversarial alignment. 
Stream 2 addresses temporal and cross-view challenges through memory-based adaptation, constructing platform-specific memories that maintain identity consistency across aerial and ground perspectives. 
Stream 3 targets the fundamental modality gap between RGB and IR imagery using anaglyph-based intermediary learning and bidirectional aggregation. 
These streams are integrated through a feature fusion module that preserves complementary information while reconciling contradictions across platforms and modalities. This multi-stream approach enables our framework to effectively disentangle the entangled variations in appearance, viewpoint, and modality that characterize aerial-ground visible-infrared person Re-ID.

\section{Hyperparameter Ablation Study}
\label{sec:hyperparameter_ablation}

We conducted ablation studies to determine the optimal values for the loss function hyperparameters ($\lambda_1$, $\lambda_2$, $\lambda_3$, and $\lambda_4$). Tab.~\ref{tab:lambda_ablation} shows the performance of our model with different hyperparameter configurations on the Aerial→Ground I2V protocol, which is one of the most challenging scenarios in our dataset.

\begin{table}[h]
\centering
\caption{Ablation study on hyperparameters $\lambda_1$, $\lambda_2$, $\lambda_3$, and $\lambda_4$ using Aerial→Ground I2V performance.}
\label{tab:lambda_ablation}
\setlength{\tabcolsep}{4pt}
\begin{tabular}{cccc|cc}
\toprule
$\lambda_1$ & $\lambda_2$ & $\lambda_3$ & $\lambda_4$ & Rank-1 & mAP \\
\midrule
0.5 & 1.0 & 0.5 & 1.0 & 15.21 & 18.43 \\
0.5 & 1.0 & 1.0 & 1.5 & 17.46 & 20.12 \\
0.5 & 1.5 & 1.0 & 1.5 & 18.32 & 21.05 \\
1.0 & 1.0 & 1.0 & 1.0 & 17.65 & 20.33 \\
1.0 & 1.5 & 0.5 & 1.5 & 18.75 & 21.42 \\
\rowcolor{gray!15}
1.0 & 1.5 & 1.0 & 1.5 & \textbf{19.83} & \textbf{22.61} \\
1.0 & 2.0 & 1.0 & 1.5 & 19.12 & 21.98 \\
1.5 & 1.5 & 1.0 & 1.5 & 18.96 & 21.75 \\
1.0 & 1.5 & 1.5 & 1.5 & 18.54 & 21.30 \\
1.0 & 1.5 & 1.0 & 2.0 & 19.07 & 21.83 \\
\bottomrule
\end{tabular}
\end{table}

As shown in Tab.~\ref{tab:lambda_ablation}, we found that setting $\lambda_1=1.0$, $\lambda_2=1.5$, $\lambda_3=1.0$, and $\lambda_4=1.5$ achieves the best performance on our most challenging retrieval scenario. The results demonstrate that:

\begin{itemize}
    \item The triplet loss ($\lambda_1$) performs best at a moderate weight of 1.0, balancing its contribution with other losses.
    \item The style attack loss ($\lambda_2$) benefits from a higher weight of 1.5, emphasizing the importance of style-invariant features for cross-platform scenarios.
    \item The cross-reconstruction loss ($\lambda_3$) is most effective at a moderate weight of 1.0, helping bridge modality gaps without overpowering other components.
    \item The video-to-memory contrastive loss ($\lambda_4$) performs optimally at 1.5, indicating the significance of temporal memory for cross-view adaptation.
\end{itemize}

We observed that increasing $\lambda_2$ and $\lambda_4$ beyond 1.5 led to performance degradation, suggesting that over-emphasizing these components can disrupt the balance between different learning objectives.

\end{document}